\documentclass[10pt,twocolumn,letterpaper]{article}

\usepackage{iccv}
\usepackage{times}
\usepackage{epsfig}
\usepackage{graphicx}
\usepackage{amsmath}
\usepackage{amssymb}
\usepackage{multirow}

\usepackage{subcaption}
\usepackage{dblfloatfix}    

\usepackage{placeins}

\usepackage[pagebackref=true,breaklinks=true,letterpaper=true,colorlinks,bookmarks=false]{hyperref}

\iccvfinalcopy 

\def\iccvPaperID{10511} 

\begin{document}

\title{Bridging the Reality Gap for Pose Estimation Networks using Sensor-Based Domain Randomization}

\author{Frederik Hagelskjær\\
SDU Robotics\\ 
University of Southern Denmark\\
Odense, Denmark\\ 
{\tt\small frhag@mmmi.sdu.dk}
\and
Anders Glent Buch\\
SDU Robotics\\
University of Southern Denmark\\
Odense, Denmark\\ 
{\tt\small anbu@mmmi.sdu.dk}
}

\maketitle

\begin{abstract}
Since the introduction of modern deep learning methods for object pose estimation, test accuracy and efficiency has increased significantly. For training, however, large amounts of annotated training data are required for good performance. While the use of synthetic training data prevents the need for manual annotation, there is currently a large performance gap between methods trained on real and synthetic data. This paper introduces a new method, which bridges this gap.

Most methods trained on synthetic data use 2D images, as domain randomization in 2D is more developed. To obtain precise poses, many of these methods perform a final refinement using 3D data. Our method integrates the 3D data into the network to increase the accuracy of the pose estimation. To allow for domain randomization in 3D, a sensor-based data augmentation has been developed. Additionally, we introduce the SparseEdge feature, which uses a wider search space during point cloud propagation to avoid relying on specific features without increasing run-time.

Experiments on three large pose estimation benchmarks show that the presented method outperforms previous methods trained on synthetic data and achieves comparable results to existing methods trained on real data.
\end{abstract}

\section{Introduction}

In this paper, we present a pose estimation method trained entirely on synthetic data. By utilizing 3D data and sensor-based domain randomization, the trained network generalizes well to real test data. The method is tested on several datasets and attains state-of-the-art performance. 

Pose estimation is generally a difficult challenge, and the set-up of new pose estimation systems is often time-consuming. A great deal of work is usually required to obtain satisfactory performance \cite{hagelskjaer2017does}. The introduction of deep learning has allowed pose estimation to obtain much better performance compared with classic methods \cite{hodan2020bop}. However, the training of deep learning methods requires large amounts of data. For new use cases, this data needs to be collected and then manually labeled. This is an extensive task and limits the usability of deep learning methods for pose estimation. The amount of manual work can be drastically reduced by generating the data synthetically.
However, getting good performance on real data with methods trained on synthetic data is a difficult task.
Classical methods generally outperform deep learning methods when using synthetic training data. 
An example of this is DPOD \cite{zakharov2019dpod}, where accuracy on the LM dataset \cite{hinterstoisser2012model} falls from 95.2\% to 66.4\%  when switching from real to synthetic training data. 
Another example is the method \cite{thalhammer2019towards} which is trained on synthetic depth data. This method only achieves a score of 46.8\%, and is outperformed by the original Linemod \cite{hinterstoisser2012model} method at 63.0\%.
As a result, most methods rely on real training data \cite{wang2019densefusion,gupta2019cullnet,hu2020single}. 

In this paper, we present a novel method for pose estimation trained entirely on synthetic data. As opposed to other deep learning methods, the pose estimation is performed in point clouds. This allows for the use of our sensor-based domain randomization, which generalizes to real data. To further increase the generalization, a modified edge feature compared to DGCNN \cite{dgcnn} is also presented. This edge feature allows for sparser and broader neighborhood searches, increasing the generalization while retaining speed. 

The trained network performs both background segmentation and keypoint voting on the point cloud. This allows the network to learn the correct object segmentation when the correct keypoint is difficult to resolve. For example, determining the correct keypoint votes for a sphere is an impossible task, while learning the segmentation is much more simple. To handle symmetry, the method allows for multiple keypoint votes at a single point. This framework allows us to test the method on three different benchmarking datasets with 55 different objects without changing any settings.
Additionally, the method is able to predict whether the object is present inside the point cloud. This makes the method able to work with or without a candidate detector method. In this article, Mask R-CNN \cite{he2017mask} is used to propose candidates, to speed up computation.



\begin{figure*}[t]
\centering
    \begin{subfigure}[t]{.195\textwidth}
      \centering
      \includegraphics[trim=250 180 180 170, clip, width=1\linewidth]{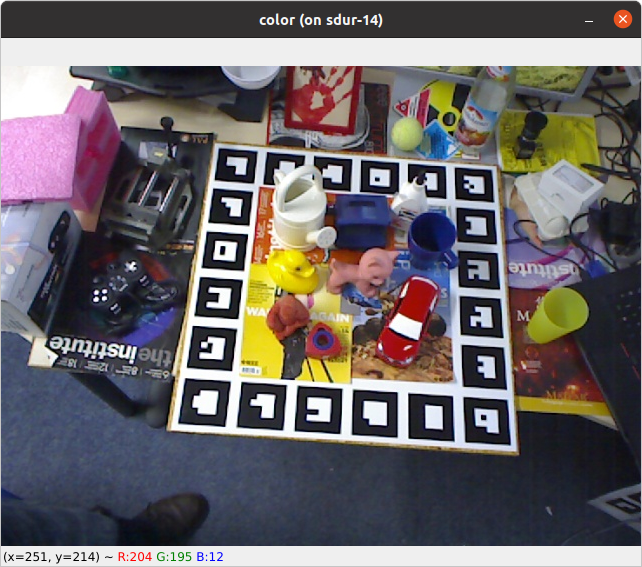}
      \caption{Initial Image.}
      \label{fig:method:sub1}
    \end{subfigure}
    \begin{subfigure}[t]{.195\textwidth}
      \centering
      \includegraphics[trim=250 180 180 170, clip, width=1\linewidth]{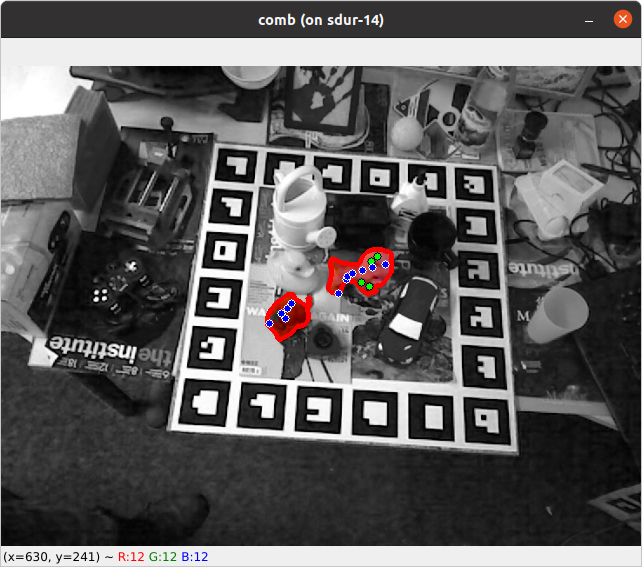}
       \caption{Mask R-CNN results \\ and cluster centers.}
      \label{fig:method:sub2}
    \end{subfigure}
    \begin{subfigure}[t]{.195\textwidth}
      \centering
      \includegraphics[trim=250 180 180 170, clip,width=1\linewidth]{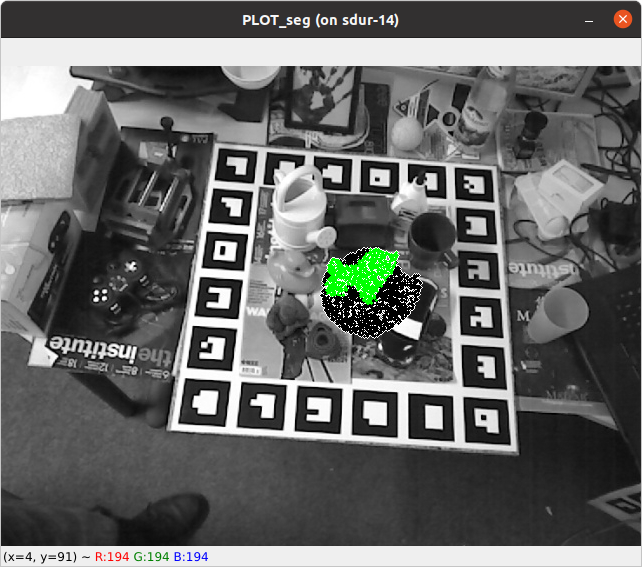}
      \caption{Background Segmentation.}
      \label{fig:method:sub3}
    \end{subfigure}
    \begin{subfigure}[t]{.195\textwidth}
      \centering
      \includegraphics[trim=250 180 180 170, clip,width=1\linewidth]{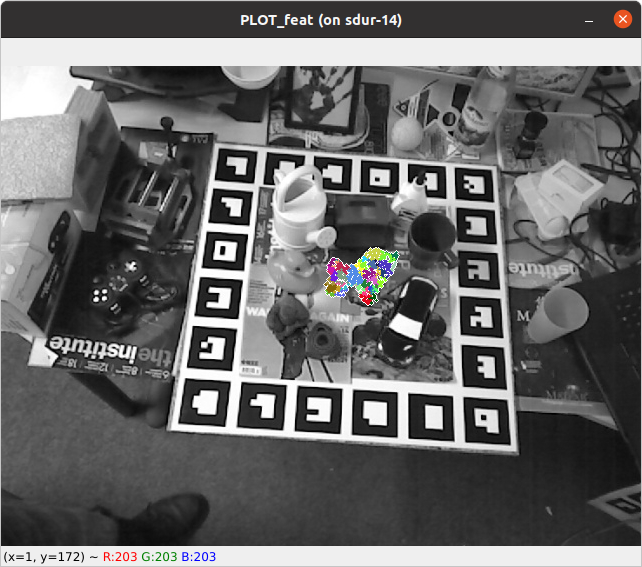}
      \caption{Keypoint Voting.}
      \label{fig:method:sub4}
    \end{subfigure}
    \begin{subfigure}[t]{.195\textwidth}
      \centering
      \includegraphics[trim=250 180 180 170, clip,width=1\linewidth]{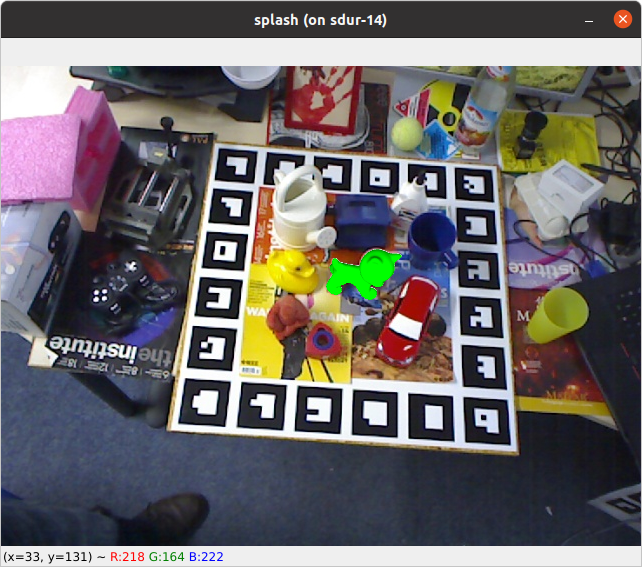}
      \caption{Final pose projected \\ into the image.}
      \label{fig:method:sub5}
    \end{subfigure}
\caption{ The pipeline of our developed method, shown in a zoomed-in view of image 10 for object 6 in the Linemod (LM) dataset. From left to right: initial image, Mask R-CNN \cite{he2017mask} and cluster detection with the four best clusters in green, background segmentation, keypoint voting, and finally the found pose in the scene shown in green. }
\label{fig:method}
\end{figure*}

Our method achieves state-of-the-art performance on the Linemod (LM) \cite{hinterstoisser2012model} dataset for methods trained with synthetic data, and outperforms most methods trained on real data. On the Occlusion (LMO) dataset \cite{brachmann2014learning} the method shows performance comparable with methods trained on real data. Additionally, on the four single instance datasets of the BOP dataset \cite{hodan2020bop}, our method outperforms all other methods trained on the same synthetic data.

The paper is structured as follows: We first review related papers in Sec.~\ref{related}. In Sec.~\ref{method}, our developed method is explained. In Sec.~\ref{evalution}, experiments to verify the method, and results are presented. Finally, in Sec.~\ref{conclusion}, a conclusion is given to this paper, and further work is discussed.




\section{Related Work}
\label{related}

Deep learning based methods have heavily dominated the performance in pose estimation for the last five years. Especially CNN-based models have shown very good performance. Several different approaches have been made to utilize CNN models for pose estimation. 

\noindent
\textbf{2D methods:}
In SSD-6D \cite{kehl2017ssd}, a network is trained to classify the appearance of an object in an image patch. By searching through the image at different scales and locations, the object can then be detected. A different approach is used in both BB-8 \cite{rad2017bb8} and another method \cite{tekin2018real} where a YOLO-like \cite{redmon2016you} network is used to predict a set of sparse keypoints. In PVNet \cite{peng2019pvnet}, the network instead locates keypoints by first segmenting the object and then letting all remaining pixels vote for keypoint locations. In PoseCNN \cite{xiang2017posecnn}, the prediction is first made for the object center, after which a regression network determines the rotation. In CDPN \cite{li2019cdpn}, the rotation and translation are also handled independently, where the translation is found by regression, and the rotation is found by determining keypoints and then applying PnP.
Similar to our method, the EPOS \cite{hodan2020epos} method uses an encoder-decoder network to predict both object segmentation and dense keypoint predictions. However, unlike our method, the network only runs in 2D images. The DPOD \cite{zakharov2019dpod} method also computes dense keypoint predictions in 2D and computes PnP, but also employs a final pose refinement.
Similar to other methods, CosyPose \cite{labbe2020cosypose} first uses an object detector to segment the image, after which a novel pose estimation based on EfficientNet-B3 \cite{tan2019efficientnet} achieves state-of-the-art performance. In addition, CosyPose can then use candidate poses from several images to find a global pose refinement. 

\noindent
\textbf{3D methods: }
In DenseFusion \cite{wang2019densefusion} initial segmentations are found in 2D, and the 2D features are then integrated with 3D features from  PointNet \cite{qi2017pointnet} before a final PnP determines the pose. Our method also employs PointNet, but unlike DenseFusion our method can perform segmentation and keypoint voting independently of 2D data. More similar to our method is PointVoteNet \cite{hagelskjaer2019pointvotenet}, which uses a single PointNet network for pose estimation. However, unlike our method, PointVoteNet combines segmentation and keypoint voting into one output and does not utilize the Edge Feature from DGCNN \cite{dgcnn}. Additionally, PointVoteNet is only trained on real data and does not employ a 2D segmentation. PVN3D \cite{he2020pvn3d} is a method that combines 2D CNN and point cloud DNN into a dense feature. Similar to our approach, keypoints are used for pose estimation. As opposed to our method, which votes for a single keypoint per point, each point votes for the position of nine keypoints. The method performs very well on the LM dataset, but does not generalize to the more challenging LMO dataset.

\noindent
\textbf{Synthetic data:}
Of the above mentioned methods only SSD-6D \cite{kehl2017ssd} and DPOD \cite{zakharov2019dpod} are trained purely on synthetic data. Data is created by combining random background images with renders. An isolated instance of the object is rendered, and this render is then overlaid on a random background image from the COCO dataset \cite{lin2014microsoft}. While this approach is simple and easy to integrate, it has certain shortcomings. As the rendered image is overlaid on a background image, light conditions and occlusions of the object will be arbitrary. Additionally, only 2D methods can be used to train on such data, as any resulting depth map would be nonsensical. For DPOD the performance gap is quite clear, as the method trained on real data achieves a performance of 95.2\% recall, while the performance drops to 66.4\% when trained on synthetic data, tested on the LM dataset \cite{hinterstoisser2012model}. For SSD-6D, the performance with synthetic data is higher at 79\%, but still far from the mid-nineties of methods trained on real data. A method \cite{thalhammer2019towards} trained on synthetic depth data also exists. The objects are placed randomly in the scene, and camera positions are chosen according to the views in the dataset. The method applies domain randomization, but in contrast to our method, it is performed in 2D. The method does not perform well, and achieves a 46.8\% recall on the LM dataset \cite{hinterstoisser2012model}.
For the BOP challenge \cite{hodan2020bop} synthetic data was created for each dataset using BlenderProc \cite{denninger2019blenderproc}. In this approach, physical-based-rendering (PBR) is performed by dropping objects randomly in a scene, and randomizing camera pose, light conditions, and object properties. This allows for more realistic noise, as shadows, occlusion, and reflections are modeled, allowing for the training of 3D-based methods. Three methods, EPOS \cite{hodan2020epos}, CDPN \cite{li2019cdpn} and CosyPose \cite{labbe2020cosypose} have been trained on this data and tested on the BOP challenge \cite{hodan2020bop}. While our method is also trained on this data, we integrate both RGB and depth data by training on point clouds.



\section{Method}
\label{method}

The goal of our method is to estimate the 6D pose of a set of known objects in a scene. The pose estimation process is often hindered by the fact that the objects are occluded, and the scenes contain high levels of clutter. This makes it challenging to construct meaningful features that can match the object in the scene to the model. When estimating a 6D pose, the object is moving in 3D space. It is, therefore, necessary to use 3D data to obtain precise pose estimates \cite{drost2017introducing}. Methods using color based deep learning methods often employ depth data at the end-stage to refine the pose. However, by employing depth in the full pose estimation pipeline, the data can be integrated into the deep learning and, as we will show, produce more accurate pose estimates.

\textbf{Pose Estimation:} On the basis of this, a method for pose estimation using deep learning in point clouds has been developed. 
The point cloud consists of both color and depth information, along with surface normals.
The pose estimation is achieved by matching points in the point cloud to keypoints in the object CAD model. This is performed using a neural network based on a modified version of DGCNN \cite{dgcnn} explained in Sec.~\ref{network}. 

The network structure is set to handle point clouds with 2048 points, as for part segmentation in DGCNN \cite{dgcnn}, so the point cloud needs to be this size. This is achieved by subsampling a point sphere around a candidate point. The point sphere size is based on the object diagonal to only include point belonging to the object, but scaled to 120\% as the candidate point is not necessarily in the object center. 
If there are more than 2048 points in the point cloud, 2048 points are randomly selected. If less than 2048 points are present, the candidate point is rejected. 
However, as the sphere radius is dependent on the object radius, smaller objects result in smaller point clouds. Therefore, to avoid rejecting these point clouds, objects with a diagonal under 120mm, allow point clouds with only 512 points, compared with the standard 2048 points.
The point cloud is given as input to the network, and the network predicts both the object's presence, the background segmentation, and keypoint voting. An example of the network output is shown in Fig.~\ref{fig:method:sub3} and Fig.~\ref{fig:method:sub4}.
As the network is able to label whether the object is present in the point cloud, the object search can be performed entirely in 3D. However, this would be computationally infeasible as a large number of spheres would have to be sub-sampled and computed through the network. The first step in the method is, therefore, a candidate detector based on Mask R-CNN \cite{he2017mask}. To improve the robustness, 16 cluster centers spread over the mask are found as potential candidates. For each candidate point, point clouds are extracted, and the network computes the probability that the object is present. Expectedly, the 2D-based Mask R-CNN also returns a number of false positives, and the 3D network is able to filter out these, as shown in Fig.~\ref{fig:method:sub2}. For the four point clouds with the highest probability, the pose estimation is performed using the background segmentation and keypoint matches. The four best is selected to increase the robustness.
RANSAC is then performed on these matches, and a coarse to fine ICP refines the position. Finally, using the CAD model, a depth image is created by rendering the object using the found pose. The generated depth image is then compared with the depth image of the test scene. The best pose for each object is thus selected based on this evaluation. The pose estimation inlier distance is set to 10mm, this value is used for both the RANSAC, the ICP, and the depth check. The coarse to fine ICP continues for three iterations down to 2.5mm distance, with 10 iterations for each level. The parameter values have been found empirically, and a further study is shown in Sec.~\ref{ablation}.


\begin{figure*}[t]
    \begin{center}
 \includegraphics[width=0.95\linewidth]{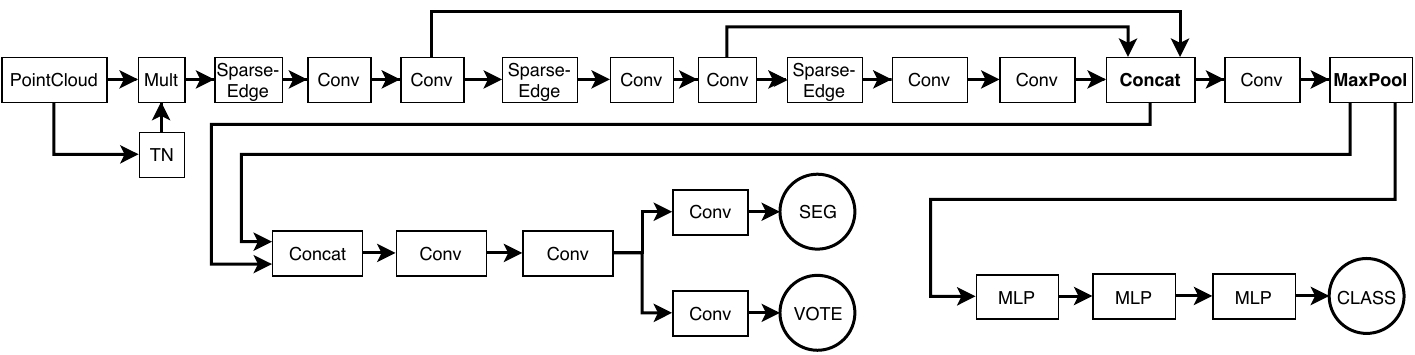}
 \caption{The structure of the neural network. The network has three outputs (shown as circles). The output CLASS is described in Sec.~3.2. The output SEG and VOTE are described in Sec.~3.3. The SparseEdge feature is described in Sec.~3.4. The \textbf{MaxPool} layer is used for the classification, while both the \textbf{MaxPool} and \textbf{Concat} layers are used for the segmentation and vote prediction. The input PointCloud is obtained from Mask R-CNN candidate detector, and the outputs are used by RANSAC to determine pose estimates. \textbf{TN} is the transform net introduced in \cite{qi2017pointnet}.}
 \label{fig:pointnet}
    \end{center}
\end{figure*}

\textbf{Set-up procedure:} The first part of the set-up procedure is the creation of keypoints. The object CAD model is sub-sampled using a voxel-grid with a spacing of 25~mm, and the remaining points are selected as keypoints. If more than 100 keypoints are present, the voxel-grid spacing is continuously increased until no more than 100 points remain.
The training data used is synthetically rendered images from the BOP challenge \cite{hodan2020bop} generated using BlenderProc \cite{denninger2019blenderproc}. The CAD model is projected into the scene, and points belonging to the object are found. The keypoints are also projected, and the nearest keypoint is found for each point. 
Point clouds are extracted from this by choosing random points and determining the label based on whether the point belongs to the object. For each image, seven point clouds are extracted, with four positive labels and three negatives. To create hard negatives for the network, one of the negative labels is found by selecting a point with a distance between 20-40~mm to the object. For each object the full process continues until 40000 point clouds have been collected for training. The network training is described in Sec.~\ref{network_training}, with the applied domain randomization described in Sec.~\ref{doma}.




\subsection{Candidate Detector}
To speed up the detection process, Mask R-CNN \cite{he2017mask} is used for an initial segmentation of the objects. The network is trained to predict an image mask of the visible surface of all objects in the scene, which we then use to get a number of candidate point clouds for the subsequent stages.

Instead of using a hard threshold for detected instances, we always return at least one top instance detection along with all other detections with a confidence above the standard threshold of 0.7. To train the network, the same synthetic data source is used, but now with image-specific randomizations. The images are randomly flipped horizontally and Gaussian blurring and noise are added with a standard deviation of, respectively 1.0 and 0.05. Additionally, hue and saturation shifts of 20 are added. Apart from this, the parameters are set according to the MASK R-CNN implementation \cite{matterport_maskrcnn_2017}. It is initialized with weights trained on the COCO dataset \cite{lin2014microsoft}, and is trained for 25 epochs. However, as the TUDL \cite{hodan2018bop} dataset only contains three objects it is trained much faster, and 50 epochs are used instead.

\subsection{Network Structure}
\label{network}

The network structure for our method is shown in Fig.~\ref{fig:pointnet}. While the network shares similarities with DGCNN \cite{dgcnn}, e.g. the size of each layer is the same, several differences exist. As opposed to DGCNN, which has a single output of either classification or segmentation, our network can output three different predictions simultaneously: point cloud classification, background segmentation and keypoint voting. 
The networks ability to perform point cloud classification and background segmentation makes it less dependent on the candidate detector. Even if false positives are presented to the network, it can filter out wrong point clouds.
As the background segmentation and keypoint votes are split into two different tasks, the network is able to learn object structure independently of the keypoint voting. This makes it easier to train the network on symmetric objects where the actual keypoint voting is difficult.


\subsection{Vote Threshold}
\label{threshold}
Before the network output of background segmentation and keypoint voting can be used with the RANSAC algorithm, they need to be converted to actual matches between scene points and object keypoints. 
The point cloud can be represented as a matrix $P$, consisting of $n$ points $p$. For each point $p_i$ in the point cloud, the network returns $s(p_i)$, representing the probability of belonging to the object vs. background. We use a lower threshold of 0.5 for classifying foreground objects.

The network also returns the keypoint vote matrix $V$ of size $n \times m, $ where $m$ is the number of keypoints in the model. For each point we then have the vector of probabilities $V(p_i)$. The highest value in $V(p_i)$ is the keypoint vote which the point $p_i$ is most likely to belong to. However, the probability distribution cannot always be expected to be unimodal. In the case of objects which appear symmetric from certain views, a point is equally likely to belong to multiple keypoints \cite{hodan2020epos}. To account for this uncertainty in our model, a scene point is allowed to vote for multiple keypoints. The approach is shown in Eq.~\ref{eqn:thre2}. For each $v_j(p_i) \in V(p_i)$ a softmax is applied and if any vote is higher than the maximum with an applied weight $\tau$, it is accepted:
\begin{equation}
\label{eqn:thre2}
    v_j(p_i) > \tau\cdot\max_{k=1}^{m}( v_k(p_i) )
\end{equation}
This allows for similar keypoints to still count in the voting process, relying on RANSAC to filter out erroneous votes. In all experiments, we use $\tau = 0.95$.


\subsection{SparseEdge Feature}
\label{sparse}

The edge feature introduced in DGCNN \cite{dgcnn} allows PointNet-like networks \cite{qi2017pointnet} to combine point-wise and local edge information through all layers. By using this edge feature, DGCNN significantly increased the performance compared to PointNet. The edge feature consists of two components, a k-NN search locating the nearest points or features, followed by a difference operator between the center point and its neighbors. The end result is a $k \times i$ feature where $k$ is the number of neighbors and $i$ is the dimension of the point representation in a layer. As the data structure from real scans is noisy, it is desirable to have a larger search space for neighbors. An increased search space will allow the method to learn a broader range of features, not only relying on the closest neighbour points. However, this increased representation capacity will also increase the computation time of the network. 

\begin{figure}[t]
    \begin{center}
 \includegraphics[width=0.80\linewidth]{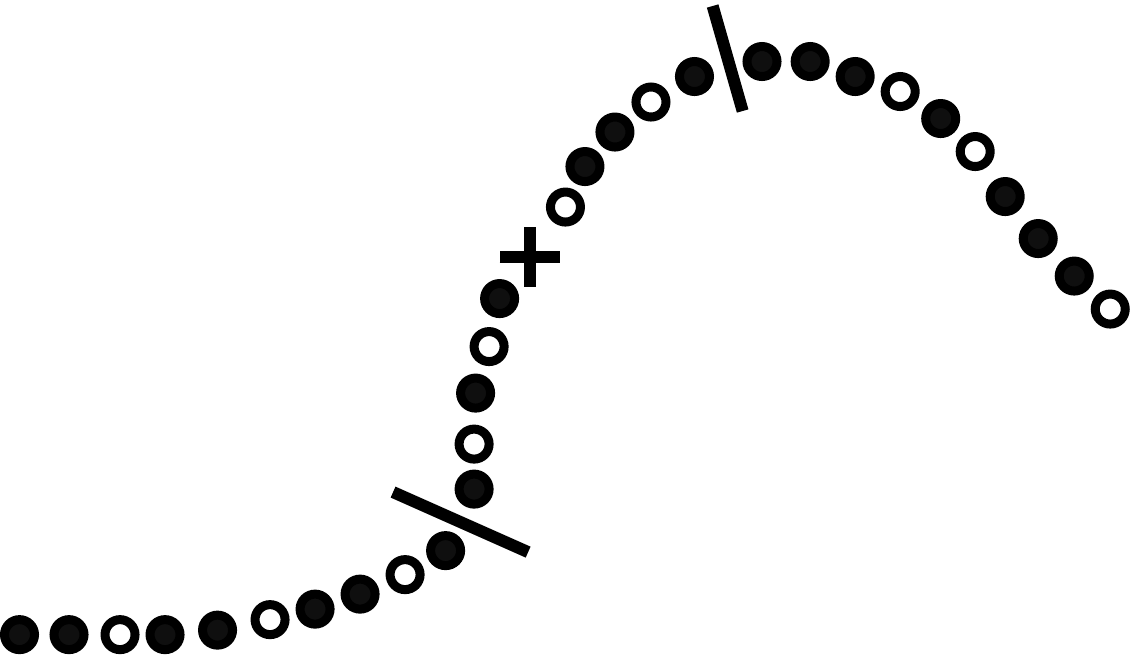}
 \caption{Example of the SparseEdge feature selection with $k=10$. The plus sign shows the center point of the k-NN. While, the standard approach select the points within the two dashes, our method select the points with white centers.}
 \label{fig:sparse}
    \end{center}
\end{figure}

To overcome this, we introduce the SparseEdge feature. The SparseEdge feature is made to maintain the performance of the edge feature, but with less run-time. Instead of selecting the $k$ nearest neighbors, a search is performed with $3k$ neighbors, and from these, a subset of $k$ is then selected. The method is shown in Fig.~\ref{fig:sparse}. At training time the $k$ neighbors are selected randomly while at test time the feature is set to select every third in the list of neighbors, sorted by the distance to the center point. The random selection at training time ensures that the network does not pick up specific features. In our experiments, $k$ is set to 10. The effectiveness of the SparseEdge is validated in Sec.~\ref{ablation}. 

\subsection{Sensor-Based Domain Randomization}
\label{doma}

\begin{figure}[t]
    \begin{center}
 \includegraphics[width=1.0\linewidth]{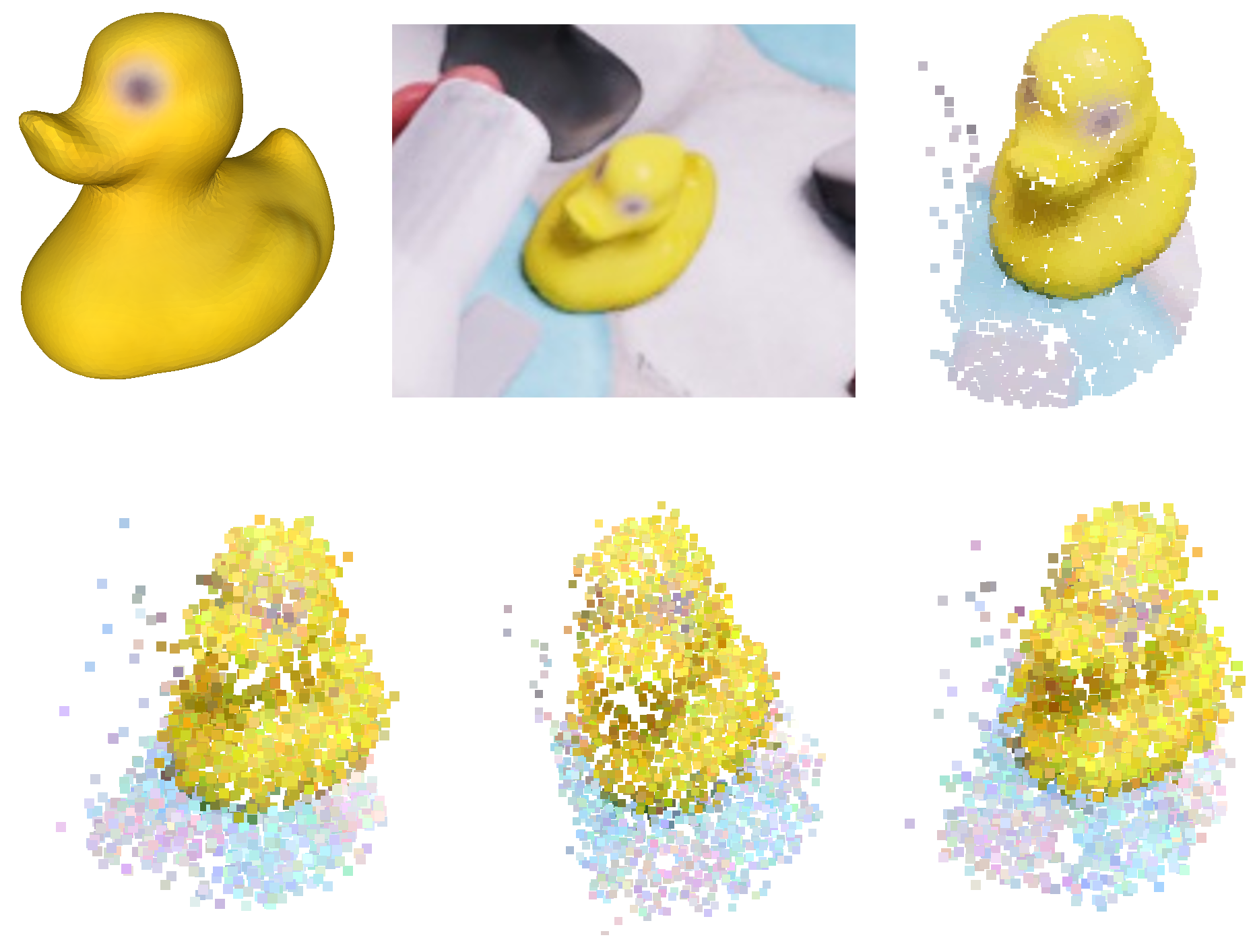}
 \caption{Example of the data sampling and domain randomization employed in this paper. From top left: CAD model of the object, object in the rendered scene, sampled point cloud, three visualizations of domain randomization applied to the point cloud.}
 \label{fig:dr_duck}
    \end{center}
\end{figure}

The synthetic training data is generated using BlenderPROC \cite{denninger2019blenderproc}. As the training data is obtained by synthetic rendering, a domain gap will exist between the training and test data. During rendering, some of the standard approaches for modeling realistic variations are included. This includes placing the objects in random positions and using random camera positions. Additionally, different types of surface material and light positions are added to the simulation, but only to the RGB part. The only disturbances on the depth part are occlusions and clutter in the simulations \cite{denninger2019blenderproc}. From the given simulated RGB-D data, we reconstruct point clouds with XYZ positions, RGB values, and estimated surface normals. The XYZ values are defined in mm, while the remaining are normalized to $[0,1]$. The standard approach for data augmentation in point clouds is a Gaussian noise with $\sigma = 0.01$ \cite{qi2017pointnet, dgcnn}. As the general approach is to normalize the point cloud size, the standard for the XYZ deviation amounts to 1\% of the point cloud size. 

For this paper the focus is on depth sensors like the Kinect with a resolution of 640x480~px. The sensor model is based on the Kinect sensor \cite{zhang2012microsoft}. Extensive analyses of the error model of the Kinect sensor have been performed \cite{nguyen2012modeling, choo2014statistical}. Modelling realistic noise is very difficult as the surface properties are unknown, and non-Lambertian reflections can cause highly non-Gaussian noise. Additionally, we face the problem that the provided CAD models do not perfectly model the 3D structure and surface texture of the objects. The goal is, therefore, not to model the noise correctly, but to model noise that gives the same error for the pose estimation. A model trained with this noise will then generalize better to the real test data. 

From the noise model one noteworthy aspect is that the error for each pixel is Gaussian and independent of its neighbors \cite{choo2014statistical}. Another important aspect is that the error depends on the angle and distance to the camera \cite{nguyen2012modeling}. The angular error is mostly insignificant when lower than $60^\circ$ and then drastically increases. The angular error is, therefore, regarded as a point dropout, and is omitted in the noise model. The noise level can, therefore, be described as Eq.~\ref{eqn:noise_depth} \cite{nguyen2012modeling}, where the constants are derived empirically.

\begin{equation}
\label{eqn:noise_depth}
   \sigma_{z}(z)=0.0012+0.0019(z-0.4)^2
\end{equation}

The distance to the objects in the datasets is between 0.3 and 2.0 meters. From Eq.~\ref{eqn:noise_depth} this gives noise levels of 1.5~mm to 6~mm. The selected $z$ distance is chosen to be 1.45 meters as this is the average maximum distance of the five tested datasets in this paper. Given $z=1.45$ the returned noise level from the formula is approximately 3~mm, which is added as Gaussian noise to the XYZ part of the point cloud.


 

Additionally, a zero-centered Gaussian noise with a $\sigma$ of 0.06 is added randomly to the color values and the normal vectors. To handle overall color differences in the CAD model texture, all RGB values in the point cloud are also shifted together with a $\sigma$ of 0.03. To increase generalization, random rotations are applied to the point clouds. These rotations are limited to $15^\circ$ so the object rotations remain towards the camera as in the real test images. As the real test background is unknown, it is desirable also to learn the object structure independently of any background. To enable this, half of point clouds with the object present have all background points removed. 

The process of sampling the training data and applying the domain randomization is shown in Fig.~\ref{fig:dr_duck}. The effect of the domain randomization is validated in Sec.~\ref{ablation}.

\subsection{Multi-Task Network Training}
\label{network_training}

As three different outputs are trained simultaneously, a weighing of the loss terms is required. The split is set according to the complexity of the different tasks, with the weights set at $w_{l}=0.12$, $w_{s}=0.22$, $w_{v}=0.66$ for point cloud label, background segmentation, and keypoint voting, respectively. An additional loss, $L_{MD}$, is added for the Transform Matrix as according to \cite{qi2017pointnet}, with weight $w_{MD} = 10^{-3}$. The full loss is shown in Eq.~\ref{eqn:losst}.

\begin{equation}
\label{eqn:losst}
   L_{total} =  w_{l} L_{l} + w_{s} L_{s} + w_{v} L_{v} + w_{MD} L_{MD} 
\end{equation}

Here $L_{l}$ is the label loss found by the cross entropy between the correct label and the softmax output of the prediction. The loss for the background segmentation $L_{s}$ is found in Eq.~\ref{eqn:losss}, where $H$ is the cross entropy, $s_{i}$ is the correct segmentation for a point, $q_{i,seg}$ is the softmax of segmentation predictions for a point, and $n$ is the number of points in the point cloud.

\begin{equation}
\label{eqn:losss}
   L_{s} = \dfrac{\sum_{i}^{n}  H(s_{i},q_{i,seg}) }{ n }
\end{equation}

When computing the keypoint voting loss, $L_{v}$, only the loss for points belonging to the object is desired. This is achieved by using $s_{i}$ which returns zero or one, depending on whether the point belongs to background or object, respectively. The loss is thus computed as in Eq.~\ref{eqn:lossf}, where  $q_{i,vote}$ is the softmax of the keypoint vote, and $v_{i}$ is the correct keypoint.
 
\begin{equation}
\label{eqn:lossf}
   L_{v} = \dfrac{\sum_{i}^{n} H(v_{i},q_{i,vote}) s_{i} }{ \sum_{i}^{n} s_{i} }
\end{equation}

The network is trained with a batch size of 48 over 40 epochs. For each object, the dataset consists of 40000 point clouds, making the complete number of training steps 1600000. The learning rate starts at 0.001 and is clipped at 0.00005, with a decay rate of 0.5 at every 337620 steps.
Batch normalization \cite{ioffe2015batch} is added to all convolutional layers in the network, with parameters set according to \cite{dgcnn}.

\section{Evaluation}
\label{evalution}
To verify the effectiveness of our developed method, and the ability to generalize to real data, we test on several benchmarking datasets. The methods compared against are all explained in Sec.~\ref{related}.
The method is tested on the popular LM \cite{hinterstoisser2012model} and LMO \cite{brachmann2014learning} datasets. As the synthetic data is obtained using the method introduced for the BOP challenge \cite{hodan2020bop}, the method is also compared with other methods using this synthetic data. The same trained weights were used to test both the LM and the LMO dataset, and the same weights were also used for the LM and LMO parts of the BOP challenge. An ablation study is also performed to verify the effect of our contributions, the sensor-based domain randomization, and the SparseEdge feature.

\begin{table}[t]
    \centering
    \scriptsize
\begin{tabular}{|l|c|c|c|c|c||c|c|c|}
\hline
  \begin{tabular}{@{}c@{}} Training \\ Data \end{tabular} & \multicolumn{5}{c||}{\textbf{Real}} & \multicolumn{3}{c|}{\textbf{Synthetic}}       \\
  \hline
  Modality & \textbf{RGB} & \multicolumn{4}{c||}{\textbf{RGB-D}} & \multicolumn{3}{c|}{\textbf{RGB-D}}       \\
  \hline
  & \cite{peng2019pvnet} & \cite{wang2019densefusion} & \cite{zakharov2019dpod} & \cite{hagelskjaer2019pointvotenet} & \cite{he2020pvn3d}  & \cite{zakharov2019dpod} & \cite{kehl2017ssd} & Ours \\
\hline
Ape      & 43.6    & 92      & 87.7       & 80.7 &  97.3  & 55.2      & 65    & 97.7 \\
Bench v. & 99.9    & 93      & 98.5       & 100  &  99.7  & 72.7      & 80    & 99.8 \\
Camera   & 86.9    & 94      & 96.1       & 100  &  99.6  & 34.8      & 78    & 98.3 \\
Can      & 95.5    & 93      & 99.7       & 99.7 &  99.5  & 83.6      & 86    & 98.8 \\
Cat      & 79.3    & 97      & 94.7       & 99.8 &  99.8  & 65.1      & 70    & 99.9 \\
Driller  & 96.4    & 87      & 98.8       & 99.9 &  99.3  & 73.3      & 73    & 99.2 \\
Duck     & 52.6    & 92      & 86.3       & 97.9 &  98.2  & 50.0      & 66    & 97.8 \\
Eggbox*  & 99.2    & 100     & 99.9       & 99.9 &  99.8  & 89.1      & 100   & 97.7 \\
Glue*    & 95.7    & 100     & 96.8       & 84.4 &  100   & 84.4      & 100   & 98.9 \\
Hole p.  & 81.9    & 92      & 86.9       & 92.8 &  99.9  & 35.4      & 49    & 94.1 \\
Iron     & 98.9    & 97      & 100        & 100  &  99.7  & 98.8      & 78    & 100  \\
Lamp     & 99.3    & 95      & 96.8       & 100  &  99.8  & 74.3      & 73    & 92.8 \\
Phone    & 92.4    & 93      & 94.7       & 96.2 &  99.5  & 47.0      & 79    & 99.1 \\
\hline
Average  & 86.3    & 94.3    & 95.15      & 96.3  & 99.4  & 66.4      & 79    &  98.0  \\
\hline
\end{tabular}

\caption{ Results for the LM dataset \cite{hinterstoisser2012model} in \% accuracy with the ADD/I score. The competing methods are DPOD \cite{zakharov2019dpod}, SSD-6D \cite{kehl2017ssd} (obtained from \cite{wang2019densefusion}), PVNet \cite{peng2019pvnet}, DenseFusion \cite{wang2019densefusion}, PointVoteNet \cite{hagelskjaer2019pointvotenet} and PVN3D \cite{he2020pvn3d}. Rotation invariant objects are marked with an *. }
\label{tab:linemod}
\end{table}

\begin{table}[t]
    \centering
    \small
    \begin{tabular}{|l|c|c|c|c|c||c|}
  \hline
   \begin{tabular}{@{}c@{}} Training \\ Data \end{tabular} & \multicolumn{5}{c||}{\textbf{Real}} & \multicolumn{1}{c|}{\textbf{Synthetic}}       \\  
 \hline
    Modality & \multicolumn{2}{c|}{\textbf{RGB}} & \multicolumn{3}{c||}{\textbf{RGB-D}} & \multicolumn{1}{c|}{\textbf{RGB-D}}       \\  
 \hline

    & \cite{xiang2017posecnn} &  \cite{peng2019pvnet} & \cite{xiang2017posecnn} & \cite{hagelskjaer2019pointvotenet} & \cite{he2020pvn3d} & Ours \\
 \hline
 Ape         & 9.60 & 15.0 &  76.2 & 70.0 & 33.9 & 66.1 \\
 Can         & 45.2 & 63.0 &  87.4 & 95.5 & 88.6 & 91.5 \\
 Cat         & 0.93 & 16.0 &  52.2 & 60.8 & 39.1 & 60.7 \\
 Driller     & 41.4 & 25.0 &  90.3 & 87.9 & 78.4 & 92.8 \\
 Duck        & 19.6 & 65.0 &  77.7 & 70.7 & 41.9 & 71.2 \\
 Eggbox*     & 22.0 & 50.0 &  72.2 & 58.7 & 80.9 & 69.7 \\
 Glue*       & 38.5 & 49.0 &  76.7 & 66.9 & 68.1 & 71.5 \\
 Hole p.     & 22.1 & 39.0 &  91.4 & 90.6 & 74.7 & 91.5 \\
 \hline
 Average     & 24.9 & 40.8 & 78.0 & 75.1 & 63.2 & 77.2  \\
 \hline
    \end{tabular}
    \caption{ Results on the LMO dataset \cite{brachmann2014learning} in \% accuracy with the ADD/I score. The score for \cite{he2020pvn3d} is from \cite{hesupplementary}. Rotation invariant objects are marked with an *.}
    \label{tab:occlusion_color}
\end{table}


\begin{table}[t]
    \centering
    \small
    \begin{tabular}{|l|c|c|c|c|c|c|}
 \hline
 & \textbf{RGB} & \textbf{RGB} & \textbf{RGB-D} & \textbf{RGB} & \textbf{D} & \textbf{RGB-D} \\
 \hline
 & \cite{hodan2020epos} & \cite{li2019cdpn} &  \cite{li2019cdpn} & \cite{labbe2020cosypose} & \cite{vidal2018method} & Ours \\
\hline
 LMO  & 54.7 & 62.4 & 63.0 & 63.3 & 58.2 & 68.4 \\
 TUDL & 55.8 & 58.8 & 79.1 & 68.5 & 87.6 & 78.2 \\
 HB   & 58.0 & 72.2 & 71.2 & 65.6 & 70.6 & 68.7 \\
 YCBV & 49.9 & 39.0 & 53.2 & 57.4 & 45.0 & 58.5 \\
 \hline
 Avg. & 54.6 & 58.1 & 66.6 & 63.7 & 65.4 & 68.2 \\
 \hline
    \end{tabular}
    ~
    \caption{Results in \% using the BOP metric for methods trained on synthetic training data on the four single instance multiple object (SiMo) datasets of the BOP 2020 challenge: LMO \cite{brachmann2014learning}, TUDL \cite{hodan2018bop}, HB \cite{kaskman2019homebreweddb}, and YCBV \cite{xiang2017posecnn}}
    \label{tab:bop}
\end{table}


\subsection{Linemod (LM) and Occlusion (LMO)}
The LM dataset \cite{hinterstoisser2012model} presents 13 objects, one object in each scene, with high levels of clutter, and some levels of occlusion. For each object, approximately 1200 images are available. The general procedure for training on the LM dataset is to use 15\% of the dataset for training, around 200 images, and test on the remaining 85\%. However, as we have trained only on synthetic data, our method is tested both using the 85\% split and using all images in the dataset; the resulting score is the same. The test results are shown in Tab.~\ref{tab:linemod}, including other recent methods trained on both real and synthetic data. Our method clearly outperforms other methods using synthetic data and outperforms most methods using real training data. 
In the LMO dataset, eight objects from the LM dataset have been annotated, many of these with very high levels of occlusion. The general procedure for testing deep learning algorithms on the LMO dataset is to use the full LM dataset for training each object, giving approximately 1200 training images for each object. Our method is the only one tested on the LMO dataset using only synthetic training. The result on the LMO dataset is shown in Tab.~\ref{tab:occlusion_color}. Our method is comparable with state-of-the-art methods using real training data. Compared with PVN3D \cite{he2020pvn3d} which achieved the highest score on the LM dataset, but low scores on the LMO dataset, our method performs well for both datasets.

Our results show that a single method trained with synthetic data, without any changes in parameters can achieve very good results in two different scenarios.

\subsection{BOP Challenge on SiMo datasets}
The synthetic training data was generated for the BOP challenge \cite{hodan2020bop}, and several other algorithms have also been trained on this data. To further validate our work, we compare it against these other methods. 

\begin{figure*}[hb]
    \centering
    \begin{subfigure}[t]{.33\textwidth}
      \centering
      \includegraphics[trim=100 0 50 70, clip, width=0.49\linewidth]{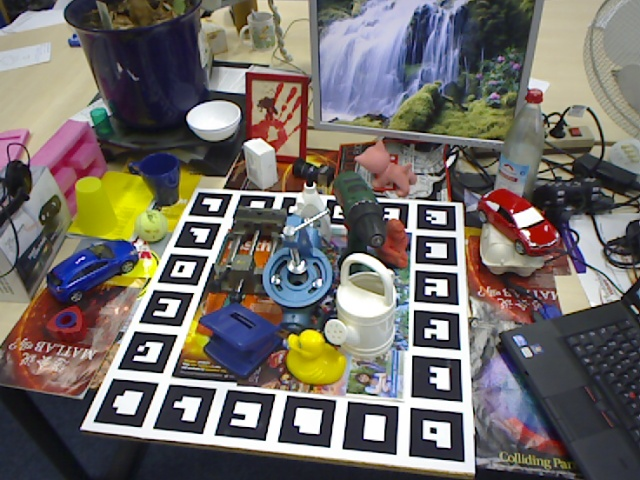}
      \includegraphics[trim=100 0 50 70, clip,width=0.49\linewidth]{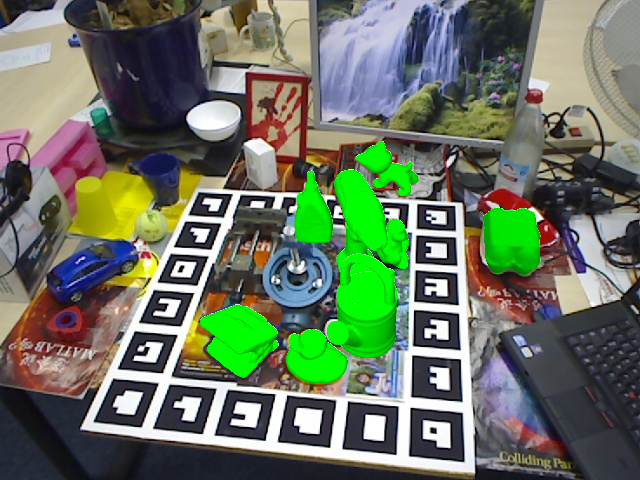}
      \caption{LMO - Scene 2 - Image 13}
      \label{fig:detections:lmo}
    \end{subfigure}%
    ~
    \begin{subfigure}[t]{.33\textwidth}
      \centering
      \includegraphics[trim=100 70 50 0, clip,width=0.49\linewidth]{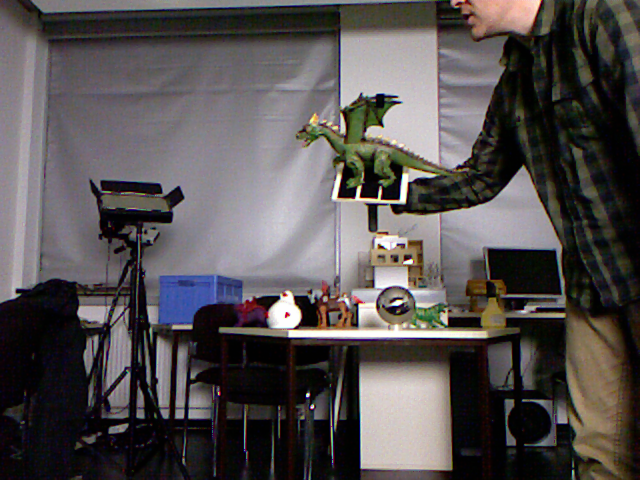}
      \includegraphics[trim=100 70 50 0, clip,width=0.49\linewidth]{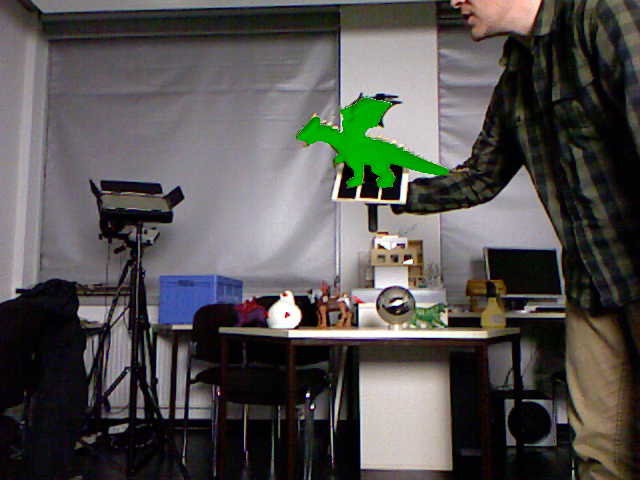}
      \caption{TUDL - Scene 1 - Image 65}
      \label{fig:detections:tudl}
    \end{subfigure}%
    ~
    \begin{subfigure}[t]{.33\textwidth}
      \centering
      \includegraphics[trim=100 0 50 70, clip,width=0.49\linewidth]{pe/bop-ycbv-test-000048001120.png}
      \includegraphics[trim=100 0 50 70, clip,width=0.49\linewidth]{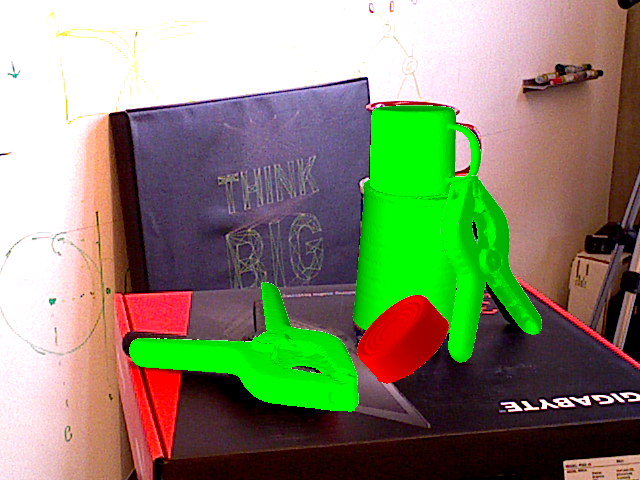}
      \caption{YCB-V - Scene 54 - Image 38}
      \label{fig:detections:ycbv}
    \end{subfigure}%
    
    
    
    \caption{ Examples of pose estimations in the BOP dataset with our method. For each image the original image is shown to the left with the pose estimation shown in right image. Successful pose estimates are shown in green and erroneous in red. }
    \label{fig:detections}
\end{figure*}

The BOP challenge consists of seven different datasets where the performance is measured for each dataset. As our method is created for single instance pose estimation, the four datasets with this configuration are retrieved, and an average is calculated. The BOP challenge score is based on an average of three metrics \cite{hodan2020bop}, making the comparison with 2D methods more equal. We use the same metric to calculate our performance. We include the results for all methods trained on the synthetic data from the competition as well as last year's winner \cite{vidal2018method}.
The results are shown in Tab.~\ref{tab:bop}. It is seen that our method is able to outperform other methods trained on the synthetic data along with last year's best-performing method. Visual examples of our pose estimation are shown for different images in the BOP benchmark in Fig.~\ref{fig:detections}. 
While the main challenge \cite{hodan2020bop} does not include the LM dataset, the associated web page contains a leaderboard\footnote{\url{https://bop.felk.cvut.cz/leaderboards/bop19_lm}} with results. Our method was tested on this dataset with the above-mentioned metric, and the resulting average BOP-specific score was 85.8\%. This outperforms the current best method \cite{zakharov2019dpod}, which has a score of 75.2\%, and is trained with real data.

\subsection{Running Time}
For a scene with a single object, the full process including pre-processing, given a 640x480 RGB-D image, takes approximately 1 second on a PC environment (an Intel i9-9820X 3.30GHz CPU and an NVIDIA GeForce RTX 2080 GPU). For the LMO dataset with eight objects in the scene the run-time is around 3.6 seconds. The time distributions for the different parts of the method is shown in Tab.~\ref{tab:time_dist}.

\begin{table}[h!]
    \centering
    {\small
    \begin{tabular}{|c|c|c|c|c|c|}
    \hline
    Part & Preproc. & \begin{tabular}{@{}c@{}} Mask \\ R-CNN \end{tabular} & DNN & \begin{tabular}{@{}c@{}} RANSAC \\ + ICP \end{tabular}  & \begin{tabular}{@{}c@{}} Depth \\ Check \end{tabular} \\ \hline
    \% Time &  15  &  8 & 24 & 49 & 4 \\ \hline
    \end{tabular}
    }
     \caption{ Percentage of time used in of our pipeline.}
    \label{tab:time_dist}
\end{table}

\noindent
\textbf{Number of Cluster Centers (CC) and number of Clusters Tested (CT):} As increasing either CC and CT will increase the run-time, a selection of the best parameter values is necessary. These are tested on the LMO dataset. In Tab.~\ref{tab:cc} CT is fixed at 4 and CC is varied. In Tab.~\ref{tab:ct} CC is fixed at 16 and CC is varied. In our implementation, the number of CC and CT is set to 16 and 4, respectively, as the optimal trade-off between performance and speed.

\begin{table}[h!]
    \centering
    {\small
    \begin{tabular}{|l|c|c|c|c|c|}
    \hline
    Cluster Centers   & 4           & 8           & 16          & 32          & 64          \\ \hline
    Run-time (s) & 2.6 & 3.0 & 3.6 & 4.8  & 7.2 \\ \hline
    Recall       & 73.8 & 76.2 & 77.2 & 77.1 & 77.3 \\ \hline
    \end{tabular}
    }
     \caption{ Recall and run-time as a result of cluster centers. }
    \label{tab:cc}
\end{table}

\begin{table}[h!]
    \centering
    {\small
    \begin{tabular}{|l|c|c|c|c|c|c|}
    \hline
    Clusters Tested   & 1           & 2           & 4          & 6          & 8    &    16         \\ \hline
    Run-time (s) & 2.2  & 2.7  & 3.6  & 4.4  & 5.7 & 8.0 \\ \hline
    Recall       & 73.9 & 76.0 & 77.2 & 77.5 & 77.4 & 76.0 \\ \hline
    \end{tabular}
     }
     \caption{ Recall and run-time as a result of clusters tested. }
    \label{tab:ct}
\end{table}

\subsection{Ablation Studies} \label{ablation}
To verify the effect of our contributions, ablation studies are performed. The test is performed by removing the contribution, retraining the network and testing against the baseline performance. The ablation studies are performed on the LMO dataset with eight objects and 1214 images, where the baseline is 77.2\% accuracy (Tab.~\ref{tab:occlusion_color}).

\noindent
\textbf{Domain randomization:} To verify the effect of our domain randomization, the network is trained with standard randomization \cite{qi2017pointnet} and without randomization. The Mask R-CNN network is the exact same for all tests. Without domain randomization the average score is 69.8\% and with standard domain randomization it is 74.4\%. The sensor-based domain randomization thus improves the performance by 11.1\% compared with no domain randomization and 3.7\% compared with standard domain randomization, both in relative numbers. If the noise level of the standard domain randomization is increased the score drops.

A more elaborated distribution of the individual parts of the ablation study is shown Tab.~\ref{tab:ablation_dr}. While the typical jitter provides some generalization, the geometric noise types (XYZ and rotation) contribute most to the generalization and are needed to achieve optimal results.

\begin{table}[h!]
    \centering
    {\small
    \begin{tabular}{|l|c|c|c|c|c|c|}
    \hline
    Removed    & None  & XYZ & Rot. & RGB & Jit. & All  \\ \hline
    Recall     & 77.2 & 73.1 & 76.1 & 77.0 & 76.9 & 69.8  \\ \hline
    \end{tabular}
    }
    \caption{ The performance on the LMO dataset for networks trained without specific Domain Randomization types.}
    \label{tab:ablation_dr}
\end{table}

\noindent
\textbf{SparseEdge feature:} Our SparseEdge method is compared with the standard edge feature from DGCNN \cite{dgcnn}, both with $k=10$ and $k=30$. For $k=10$ the score is 75.4\% and the run-time is 3.4s. For $k=30$ run-time rises to 4.1s while the score goes up to 76.9\%. For our method the run-time is 3.6s with a relative 2.4\% better performance than $k=10$ and the score is still higher than when using $k=30$. The increased performance of the SparseEdge could indicate that a higher generalization is obtained.

\section{Conclusion}
\label{conclusion}

We presented a novel method for pose estimation trained on synthetic data. The method finds keypoint matches in 3D point clouds and uses our novel SparseEdge feature. Combined with our sensor-based domain randomization, the method outperforms previous methods using purely synthetic training data and achieves state-of-the-art performance on a range of benchmarks. An ablation study shows the significance of our contributions to the performance of the method. 


For future work, instance segmentation can be added to the point cloud network. This, along with training a single network to predict keypoint votes for multiple objects, will allow us to pass an entire scene point cloud through the network for a single pass pose estimation of multiple objects.

\noindent
\textbf{Acknowledgements} The authors gratefully acknowledge the support from Innovation Fund Denmark through the project MADE FAST.



{\small
\bibliographystyle{ieee_fullname}
\bibliography{main}
}

\end{document}